\documentclass[runningheads]{llncs}
\usepackage{graphicx}
\usepackage{amsmath,amssymb} 
\usepackage{color}
\usepackage{times}
\usepackage{epsfig}
\usepackage{graphicx}
\usepackage{subfigure}
\usepackage{booktabs}
\usepackage{multirow}
\usepackage{hyperref}
\usepackage[symbol]{footmisc}
\renewcommand{\thefootnote}{\fnsymbol{footnote}}
\newcommand{\repeatthanks}{\textsuperscript{\thefootnote}}
\begin{document}
\pagestyle{headings}
\mainmatter

\def\ACCV20SubNumber{19}  

\title{Imbalance Robust Softmax for Deep Embeeding Learning} 
\titlerunning{IR-Softmax}
%
\author{Hao Zhu\inst{1}\thanks{indicates equal contribution.} \and
Yang Yuan\inst{2}\repeatthanks \and
Guosheng Hu\inst{2,4} \and Xiang Wu\inst{3} \and 
Neil Robertson\inst{4}}
\authorrunning{H. Zhu et al.}
%
\institute{Australian National University, Canberra, Australia \\\and
Anyvision, Belfast, United Kingdom
\\
\and
Reconova, Beijing, China
\and Queens University of Belfast, Belfast, United Kingdom\\\inst{1}\email{Hao.Zhu@anu.edu.au},\inst{2}\email{bengouawu@gmail.com}}
\maketitle

\begin{abstract}
Deep embedding learning is expected to learn a metric space in which features have smaller maximal intra-class distance than minimal inter-class distance.
In recent years, one research focus is to solve the open-set problem by discriminative deep embedding learning in the field of face recognition (FR) and person re-identification (re-ID).
Apart from open-set problem, we find that imbalanced training data  is another main factor causing the performance degradation of FR and re-ID, and data imbalance widely exists in the real applications. 
However, very little research explores why and how data imbalance influences the performance of FR and re-ID with softmax or its variants.
In this work, we deeply investigate data imbalance in the perspective of neural network optimisation and feature distribution about softmax. %
We find one main reason of performance degradation caused by data imbalance is that the weights (from the penultimate fully-connected layer) are far from their class centers in feature space. 
Based on this investigation, we propose a unified framework, Imbalance-Robust Softmax (IR-Softmax), which can simultaneously solve the open-set problem and reduce the influence of data imbalance. %
IR-Softmax can generalise to any softmax and its variants (which are discriminative for open-set problem) by directly setting the weights as their class centers, naturally solving the data imbalance problem. In this work, we explicitly re-formulate two discriminative softmax (A-Softmax and AM-Softmax) under the framework of IR-Softmax. We conduct extensive experiments on FR databases (LFW, MegaFace) and re-ID database (Market-1501, Duke), and IR-Softmax outperforms many state-of-the-art methods.
\end{abstract}
\section{Introduction}

Recently, convolutional neural networks (CNNs) have significantly boosted the state-of-the-art performance in many computer vision tasks especially in image classification \cite{krizhevsky2012imagenet,sermanet2013overfeat,szegedy2015going,he2015deep,he2015delving,simonyan2014very}. 
Not surprisingly, CNNs have achieved great success in the field of biometrics, in particular, face recognition (FR) \cite{taigman2014deepface,sun2014cvpr,sun2014deep,schroff2015facenet} and  person re-identification (re-ID) \cite{li2014deepreid,zheng2016person,zheng2017unlabeled}. 
This success is derived from the fact that  CNNs are able to  encode images into rich, semantic and discriminative  representations (features) which can be used to effectively measure the similarity between two identity-related images.
These two tasks (FR and re-ID) differ from general image classification in terms of two challenges: open-set setting and data imbalance in the training set. 

Open-set setting is much more widely applied than close-set for FR and re-ID. 
For open-set setting, the identities of test set are disjoint with those of training set. 
In the real world, FR and re-ID system train the CNN (feature extractor) using images collected from one specific group of people, e.g. celebrities from IMDb in CASIA WebFace ~\cite{DBLP:journals/corr/YiLLL14a} database.  
During test stage, however, the FR and re-ID systems work in places, such as one police station, where the gallery  (blacklist) and the probe  (people appear in this police station) are mostly likely disjoint with training set (e.g. those celebrities). 
In contrast, classical image classification (e.g. ImageNet  Challenge) uses the  close-set setting 
where training and test sets share the same classes.
Traditionally, both open-set and close-set problems adopt the softmax function because of its simplicity and probabilistic interpretation. Together with the cross-entropy loss, they form arguably one of the most commonly-used components in CNN architectures.

Under open-set setting, however, softmax suffers from one drawback:  deep learning with softmax loss only learns separable features that are not discriminative enough for `unseen' classes in testing. It results from the fact that softmax loss does not explicitly optimise the intra-  and inter-class distances.  To address this, some methods combine the softmax loss with metric learning \cite{sun2014deep,sun2015deeply,schroff2015facenet} to enhance the discrimination power of features. Metric learning based methods commonly suffer from the way of  building mini-batches by sampling. Other methods try to add new constraints (e.g. center loss \cite{wen2016discriminative}, large-margin term \cite{liu2016large,liu2017sphereface}, L2 normalization \cite{DBLP:journals/corr/WangXCY17,2017arXiv170309507R}) that make features more compact and thus more discriminative.

\begin{figure*}
\centering
\subfigure[WebFace-CASIA]{
\includegraphics[width=0.45\linewidth]{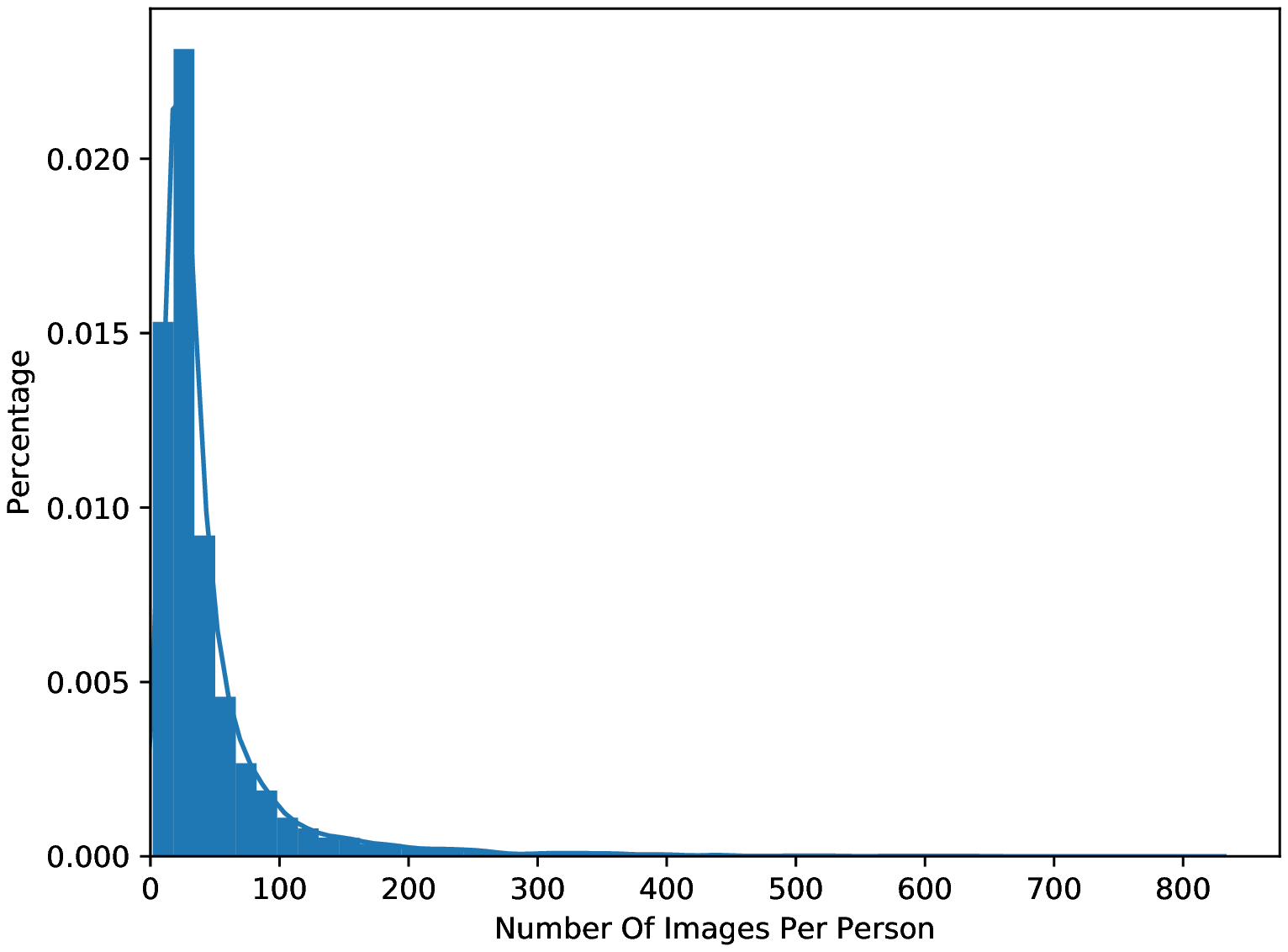}
}
\subfigure[Market-1501]{
\includegraphics[width=0.45\linewidth]{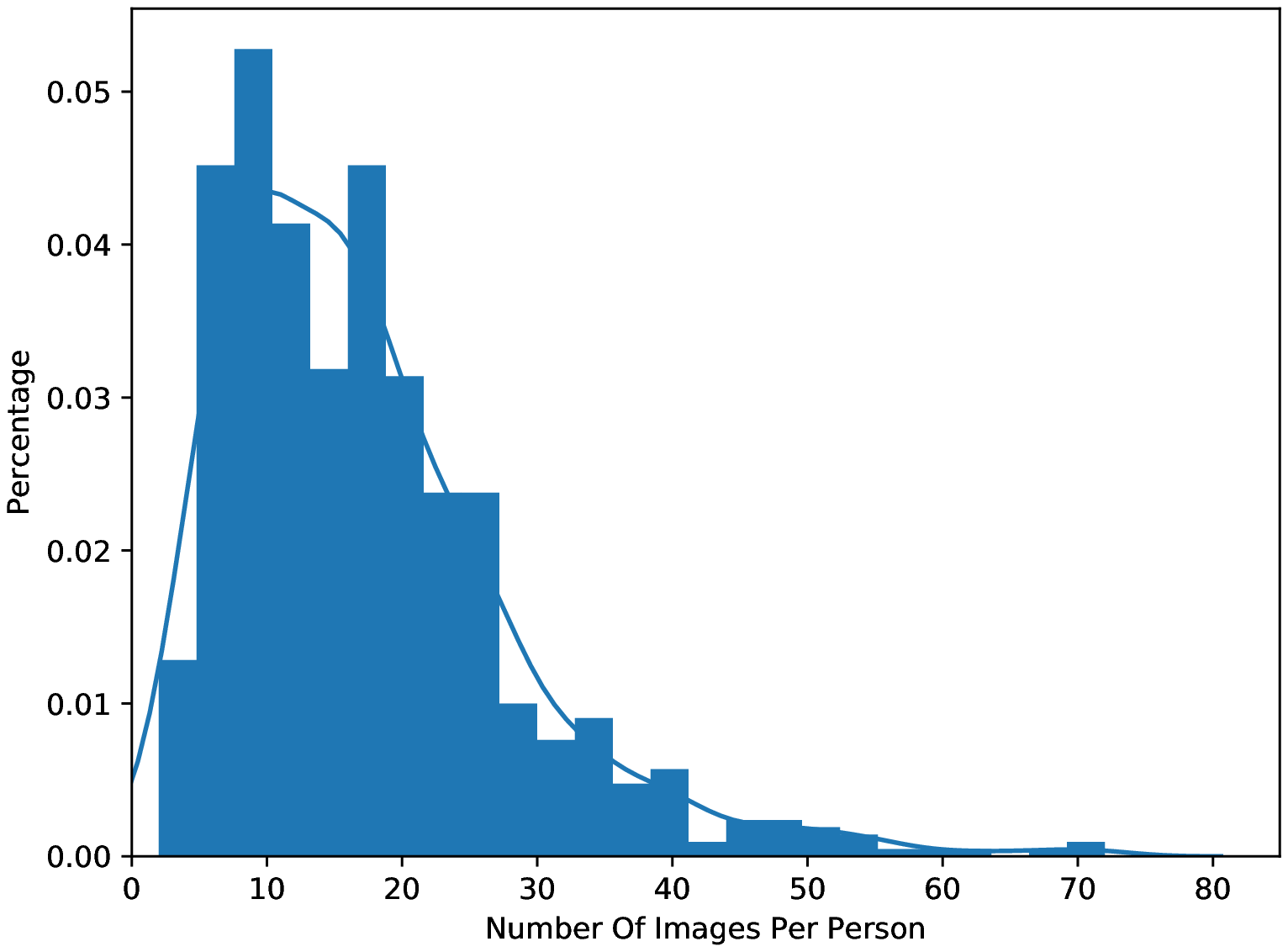}
}
\caption{Long-tailed distribution on WebFace-CASIA \cite{DBLP:journals/corr/YiLLL14a} and Market-1501 \cite{zheng2015scalable} database. The number of images per person drops drastically, and only a few identities have a large number of images.}
\label{fig:imbalance-dis}
\end{figure*}

{
Data imbalance is another challenge for FR and re-ID.  Unlike those popular datasets  MNIST \cite{lecun1998mnist}, CIFAR-10 \cite{krizhevsky2009learning} and ImageNet \cite{deng2009imagenet}, FR and re-ID datasets are commonly highly imbalanced.  
As shown in Fig.\ref{fig:imbalance-dis}, only a limited number of identities appear frequently (more than hundreds), while most of the others appear relatively rarely (fewer than ten times) in the popular face database CASIA-Webface~\cite{DBLP:journals/corr/YiLLL14a}} and re-ID database Market-1501~\cite{zheng2015scalable}. 
Surprisingly, very little research explores the problem of data imbalance in FR and re-ID. 
In this paper, we show that deep embedding learning with the most widely used softmax~(and its variants such as {A-Softmax \cite{liu2017sphereface}}) encounters difficulty in the presence of  imbalanced training data even using either metric learning or other regularizations. 
Although some softmax variants such as A-Softmax \cite{liu2017sphereface} can solve the open-set problem by learning compact features, they do not perform well when the training data is imbalanced. 
To our knowledge, the only work exploring the data imbalance problem for FR is the range loss \cite{DBLP:journals/corr/ZhangFWL016}. However, range loss does not deeply investigate the reason why this imbalance impacts the deep embedding learning.

In this work, we aim to learn deep embeddings which can achieve two targets: 1) being discriminative for open-set  and 2) being robust to data imbalance. As existing works \cite{wen2016discriminative,liu2016large,liu2017sphereface,DBLP:journals/corr/WangXCY17,2017arXiv170309507R}, target 1) can be achieved by learning compact features (i.e. reduce intra-class variance). 
To achieve target 2), we have to first investigate why data imbalance influences the performance of softmax-based deep classification. 
In this work, we explore the reason. \emph{During the back-propagation training with imbalanced data, two strengths, which determine the update of the weights (usually the penultimate fully-connected layer), are imbalanced (see Eq. \ref{equ:softmax-gradient}): the one  keeping the weights  at  their class centers is much smaller than that pushing them away. This imbalance causes the weight of the class with minor samples being far away from its class center, leading to degraded classification performance.}  Based on this analysis, target 2) can be achieved by making the weight from the class with minor samples close to its class center. 

To simultaneously achieve targets 1) and 2) , we propose a uniformed framework, Imbalance-Robust Softmax (IR-Softmax). First, IR-Softmax solves the open-set problem by being compatible with the softmax variants ( e.g. A-Softmax \cite{liu2017sphereface}, AM-Softmax \cite{2018arXiv180105599W} which can learn discriminative embeddings.
Second, motivated by the aforementioned analysis on data imbalance, IR-Softmax alleviates the influence of data imbalance by setting the weights as their class centers in the feature space instead of updating with back-propagation.
In this way, IR-Softmax effectively avoids the shift between the weights and their centers, which is the main reason of performance degradation caused by data imbalance detailed in Section \ref{s:softmax}. 
   
Our contributions can be summarised as:
   
1.   We deeply investigate the reason why data imbalance degrades the performance of softmax-based classifications in the perspective of  neural network optimisation (Section \ref{s:softmax}) and feature distribution (Section \ref{sec:irs}). 

2. IR-Softmax can  learn embeddings which are discriminative under open-set protocol. 
In particular, IR-Softmax is a unified framework,   e.g. it  can generalise to softmax and its variants (e.g. A-Softmax \cite{liu2017sphereface}, AM-Softmax \cite{2018arXiv180105599W}) to achieve discriminative feature learning. More importantly, IR-Softmax  can effectively reduce the influence of data imbalance by bridging the gap between weights (the penultimate fully-connected layer) and their class centers in feature space. 

3. Our extensive experiments demonstrate the effectiveness and generalisation of the proposed IR-Softmax, and we achieve state-of-the-art performance on challenging FR (LFW \cite{huang2007labeled}, MegaFace \cite{kemelmacher2016megaface}) and re-ID (Market-1501 \cite{zheng2015scalable},  DUKE-MTMC \cite{ristani2016performance}) benchmarks. The code will be made publicly available.

\begin{figure*}
\centering
\subfigure[Softmax, bal.]{
\includegraphics[width=0.23\linewidth]{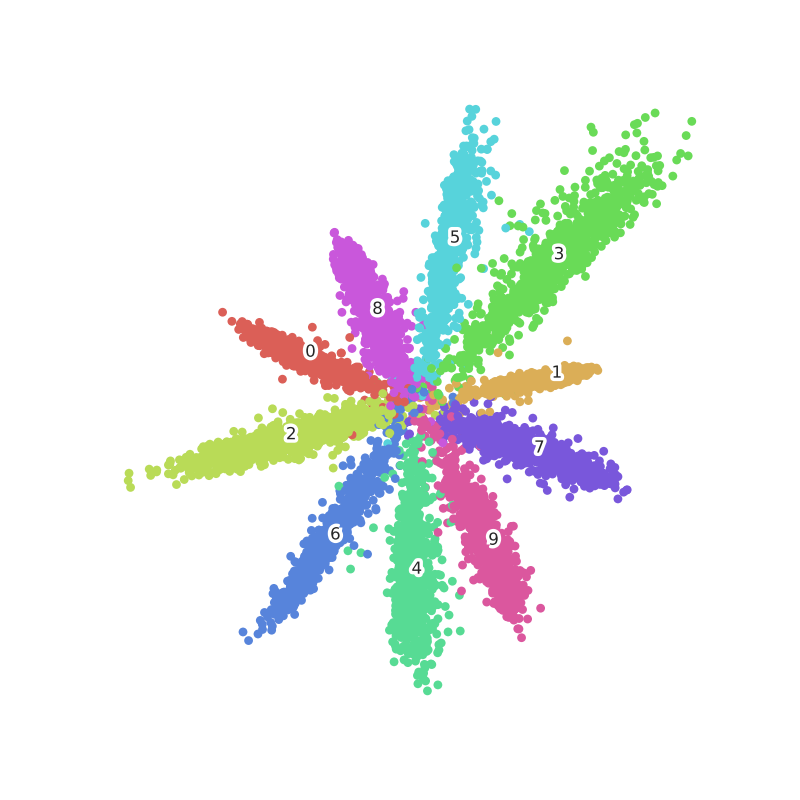}
\label{fig:softmax}
}
\centering
\subfigure[Softmax, imbal.]{
\includegraphics[width=0.23\linewidth]{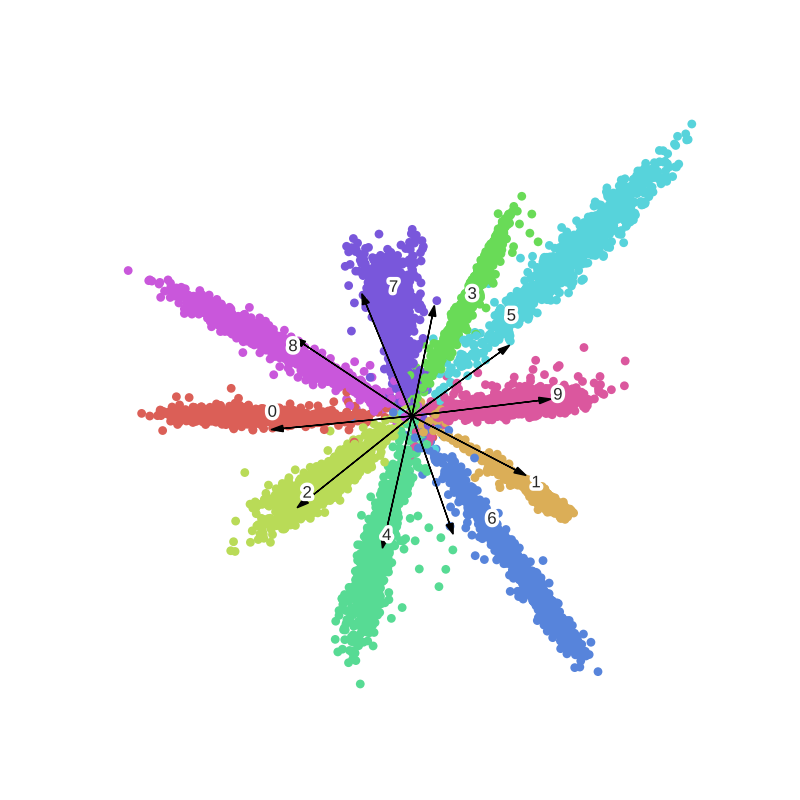}
\label{fig:Isoftmax}
}
\subfigure[A-Softmax, imbal.]{
\includegraphics[width=0.23\linewidth]{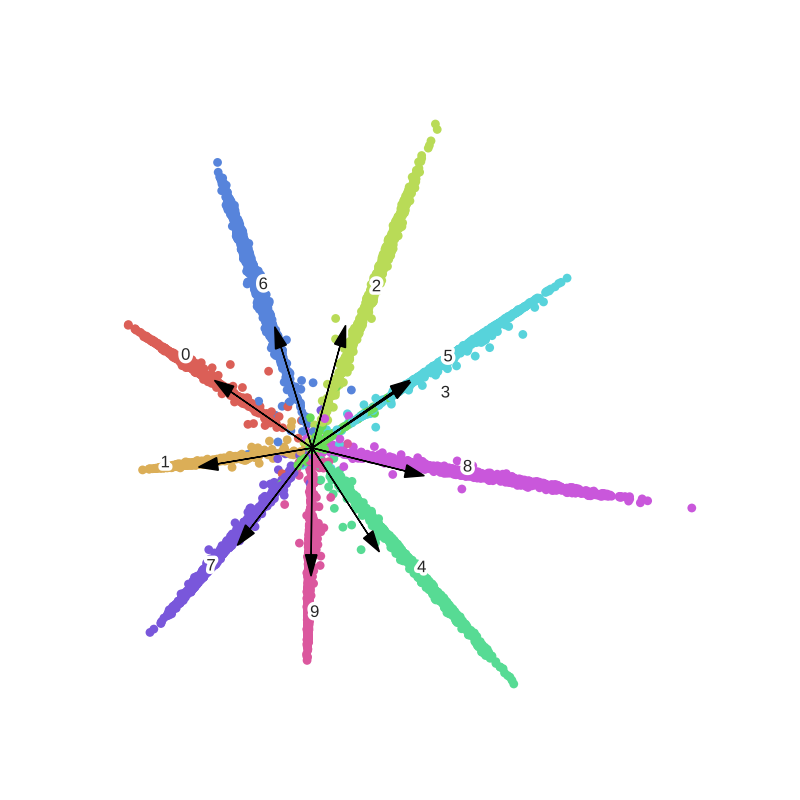}
\label{fig:ILM}
}
\subfigure[\textbf{IR-Softmax}, imbal.]{
\includegraphics[width=0.23\linewidth]{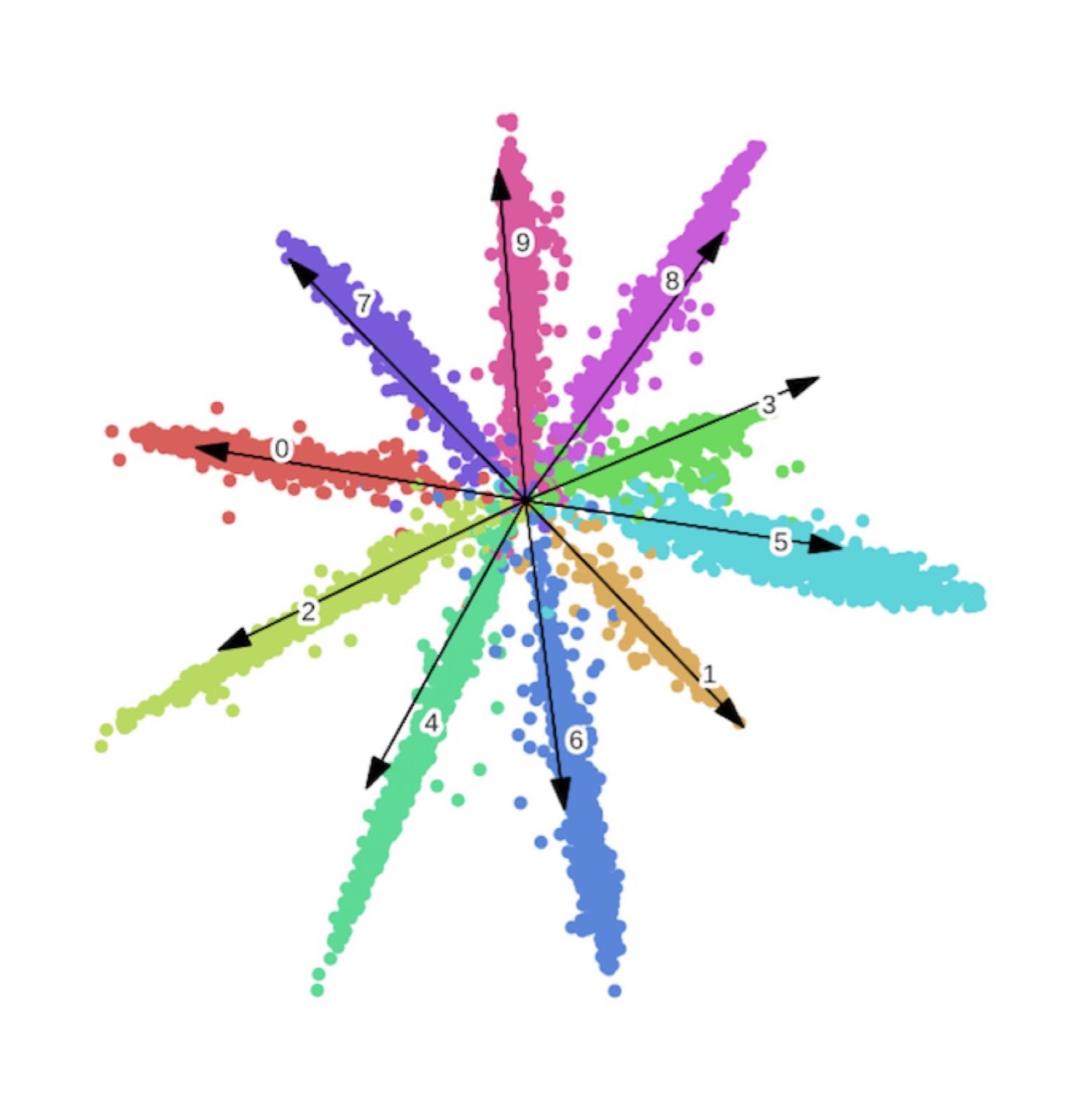}
\label{fig:ours}
}
\caption{The distribution of deeply learned features. `bal.' (balanced) setting contains 10 classes, which all has 6,000  images  from MNIST \cite{lecun1998mnist}. In contrast, `imbal.' (imbalanced) setting contains 6000 images for all classes but class `3' with  300 images.
A-Softmax in (c) refers to \cite{liu2017sphereface}.
The label of each class is plotted on its  center. 
In addition, we also plot the {weights (from the fully-connected penultimate layer)} to each class with an arrow in  (b)-(d). 
Note that our fully-connected layer consists of only 2 neurons to facilitate visualisation. }
\label{fig:fig1}
\end{figure*}

\section{Related Work}
In this section, we briefly review the methods of  discriminative feature learning in the field of face recognition (FR) and person re-identification (re-ID). Recently, two popular ways of deep embedding learning are: (1) metric learning and (2) discriminative softmax (softmax's variants which are more discriminative for open-set problem). Apart from these two strategies, we discuss the training data imbalanced problem in the field of FR and re-ID. 

\vspace{0.0cm}\noindent\textbf{Metric Learning}\quad 
Metric learning is widely used for FR and re-ID. 
In practice, to learn more discriminative features,  many works  combine softmax loss and deep metric learning loss (contrastive \cite{sun2014deep,DBLP:journals/corr/GengWXT16} loss or triplet loss \cite{schroff2015facenet}). Unlike softmax, contrastive and triplet losses accept image pairs or triplets (3 or a multiple of 3) as input respectively. For contrastive loss, 
if the input pair belongs to the same class, their features are required to be as similar as possible. Otherwise, the contrastive loss would require their distance larger than a particular margin. Similar to contrastive loss,  the triplet loss \cite{schroff2015facenet} encourages a similar distance constraint. Specifically, the triplet loss minimises the distance between an anchor sample and a positive sample (of the same identity) and maximises the distance between the anchor sample and a negative sample (of different identity). 
Clearly, the contrastive and triplet losses can 
encourage intra-class compactness and inter-class separability, making the learned feature more discriminative. However, 
both  contrastive and triplet losses  require a carefully-designed pair/triplet selection procedure. For example, using contrastive loss, it is hard to build training pairs from a mini-batch, especially for the training set with many classes. Normally the mini-batch size is not more than 256, while the number of categories is far more than 256 in the application of FR and re-ID. Clearly the online selection only produces a few positive  pairs and much more negative  ones. 

\vspace{0.0cm}\noindent\textbf{Discriminative Softmax}\quad  
Apart from metric learning, some  softmax variants are proposed, aiming to learn more discriminative features to solve the open-set problem. 
Wen et al.\cite{wen2016discriminative} add a new supervision signal, called center loss, to softmax loss for  face recognition. Specifically, the center loss simultaneously learns a feature center for each identity and penalises the distances between the deep features of examples and their corresponding feature centers. With the joint supervision of softmax loss and center loss, this method can easily obtain inter-class dispersion and intra-class compactness. Large-Margin Softmax loss \cite{liu2016large} proposes a new perspective to softmax and optimises the angles between weights and features. However, the magnitude of weights are also considered, and thus it is also sensitive to data imbalance just the same as softmax. By contrast, A-Softmax loss \cite{liu2017sphereface} controls the magnitude of weights (i.e. $\|w\|_2=1$) and thus make the weights optimised in an angular space. Although A-Softmax is theoretically  suitable for deep embedding learning, it actually does not work well in the setting of data imbalance detailed in Section \ref{sec:method}. \cite{wang2018cosface,2018arXiv180105599W,deng2019arcface} relax the margin with more efficient and effective ways. Some works \cite{DBLP:journals/corr/WangXCY17,2017arXiv170309507R,deng2019arcface}  try to optimise the features on a hyper-sphere to make features more discriminative.

\vspace{0.0cm}\noindent\textbf{Training Data Imbalance}\quad 
The aforementioned methods  ignore the problem of training data imbalance which widely exist in FR and re-ID. In \cite{ouyang2016factors}, researchers investigate many factors that influence the performance of fine-tuning for object detection with long-tailed distributions of samples. Their analysis and empirical results indicate that classes with more samples will achieve greater impact on the feature learning, and it is better to make the sample number more uniform across classes. 
In the field of FR and re-ID, unfortunately, the data imbalance problem  is much worse than object detection \cite{ouyang2016factors}. 
Specifically, few identities have more than 1000 images and many identities have fewer than 10 images. Commonly a large-scale face dataset has more than 10,000 identities \cite{DBLP:journals/corr/YiLLL14a}. However, we still cannot simply discard these identities that only have few images. 
For face recognition, identities with few images cannot provide enough intra-class information for the model, but provide inter-class information which is more useful to open-set protocol. 
Many methods \cite{DBLP:journals/corr/ZhangFWL016,shi2019docface+,zhong2019unequal,khan2019striking,yin2019feature} have been proposed to solve the data imbalance in face recognition. However,these works do not deeply investigate the reason why the imbalance impacts softmax based deep embedding learning.


\section{Methodology}
\label{sec:method}

In this section, we first provide insights into the influence of data imbalance on CNN performance by training a LeNet \cite{lecun1998mnist} on an imbalanced MNIST. 
Based on the conclusion drawn from the experiments, we propose a new loss function,  Imbalance Robust Softmax (IR-Softmax), to reduce the influence of data imbalance while perform discriminative feature learning. 
Last, we discuss the relations between the proposed method and metric learning. 

\subsection{Motivation}
\label{s:softmax}

Softmax regression (or multinomial logistic regression) is a generalisation of logistic regression to multi-class problem, therefore, softmax  can  handle $y_i \in \{1,\ldots,K\}$ (where $K$ is the number of classes). Given a training set $\{ (\mathbf{x}_i, y_i), \ldots, (\mathbf{x}_n, y_n) \}$, we  learn an embedding/projection $f(\mathbf{x})$, with which the  softmax  can be written as
\begin{equation}
J = -\frac{1}{n}\left[ \sum_{i=1}^{n} \log \frac{\exp{(f_{y_i}(\mathbf{x}_i))}}{\sum_{j=1}^K \exp(f_j(\mathbf{x}_i))}\right]
\label{equ:softmax}
\end{equation}
where $f_j$ denotes the $j$-th dimension of the learned function $f(x)$, and $n$ is the number of training samples. In CNNs, $f$ is usually the output of a fully connected layer $\mathbf{W}=[\mathbf{w}_1,...,\mathbf{w}_k]$ , so $f_j = \mathbf{w}_j^T \mathbf{x}_i + b_j$ and $f_{y_i} =\mathbf{w}_{y_i}^T\mathbf{x}_i+b_{y_i}$. 

To analyse the influence of data imbalance, we come to the neural network optimisation process (we omit the bias term for simplicity):
\begin{equation}
\begin{split}
\nabla_{\mathbf{w}_k} J &= - \frac{1}{n}\sum_{i=1}^{n}{ \mathbf{x}_i \left( 1\{ y_i = k\}  - P(k| \mathbf{x}_i) \right)}\\
&=\frac{1}{n}
(
\underbrace{\sum_{i=1}^{n}\mathbf{x}_i(P(k| \mathbf{x}_i)-1)1\{ y_i = k\}}_{term~1} + \underbrace{\sum_{i=1}^{n}\mathbf{x}_iP(k | \mathbf{x}_i)1\{ y_i \neq k\}}_{term~2})
\end{split}
\label{equ:softmax-gradient}
\end{equation}
where $P(k|\mathbf{x}_i)=\frac{\exp{(f_{k}(\mathbf{x}_i))}}{\sum_{j=1}^K \exp(f_j(\mathbf{x}_i))}$, and $1\{\cdot\}$ is the indicator function: $1\{{true}\} = 1$, and $1\{{false}\} = 0$. It can be observed that the gradient of the parameter $\nabla_{\mathbf{w}_k} J$ contains { two terms: term 1 (which is activated  if $y_i=k$) and term 2 ( if $y_i \neq k$).} 
Thus the update of parameter $\mathbf{w}_k$ during optimisation depends on the samples not only  from the $k$-th class but also  from the other classes. 
{ Term 1 is actually the weighted center of the observed class; Term 2 can be viewed as the weighted centers of all the other classes if $n$ is big enough. The update of $\mathbf{w}_k$ is determined by the balance of two strengths: one leads $\mathbf{w}_k$ to the center of class $k$ (term 1), one `pushes'  $w_k$ away from class $k$ (weighted center of all the other classes). 
If the training data is imbalanced, the update of  $\mathbf{w}_k$ corresponding to class $k$, which has much fewer samples than other classes, is fully dominated by term 2, making $\mathbf{w}_k$ being far away from center of class $k$. }

\begin{figure*}
\centering
\includegraphics[width=0.5\linewidth]{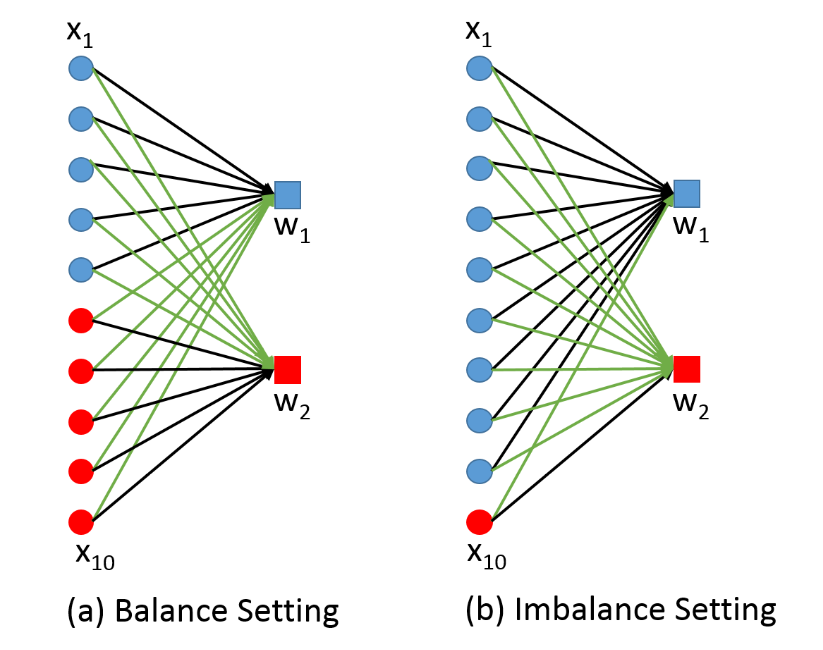}
\caption{One example of binary classification with softmax. (a) balance setting with five samples of  class 1 (blue spots) and five samples of  class 2 (red spots). (b) imbalance setting with nine samples of  class 1 and one sample of  class 2. 
Black and green lines  indicate term 1 and  term 2 in Eq.~(\ref{equ:softmax-gradient}), influencing the update of 
 $\mathbf{w}_1$ and $\mathbf{w}_2$    respectively.}
\label{fig:binary}
\end{figure*}

{
To further analyse the influence of data imbalance on optimisation, 
we take one binary classification with softmax for example.
As shown in Fig.\ref{fig:binary}(a) and (b), there are both five samples for class 1 (blue points) and class 2 (red points) for balance setting; and nine samples for class 1 and one sample for class 2 for imbalance setting. Clearly, both $\mathbf{w}_1$  and $\mathbf{w}_2$  are influenced by all the samples from class 1 and 2. 
In imbalance setting (Fig.\ref{fig:binary}(b)),  
 $\mathbf{w}_2$ is  determined by term 1 (1 sample from class 2) and term 2 (9 samples of class 1), where term 1 and 2 are detailed in Eq. (\ref{equ:softmax-gradient}). Clearly, the update of  $\mathbf{w}_2$ is dominated by term 2, which pushes $\mathbf{w}_2$ far away from the center of class 2.
}

To explicitly show the influence of data imbalance on classification performance, we conduct a toy experiment on MNIST \cite{lecun1998mnist}.
From Fig.\ref{fig:Isoftmax} and \ref{fig:ILM}, not surprisingly, the data imbalance  degrades the performance of the models trained with softmax and A-Softmax \cite{liu2017sphereface}. 
We can find the main issue caused by data imbalance: centers of relevant feature distributions being away from their weights (from penultimate fully-connected layer). 
For example, in Fig. \ref{fig:Isoftmax}, the feature center of class `3' (minor training data) and centers of  `5' and `7' (the neighbours of `3') are all distant from their weights. 
Thus these biases (feature centers being far from its weights) caused by data imbalance will induce classification error for the corresponding categories. 
Though A-Softmax can learn discriminative features, 
it suffers  from the same aforementioned bias problem as shown in Fig.~\ref{fig:ILM}. 
This  observation provides the cue to solve the data imbalance problem and inspires our solution (Fig.~\ref{fig:ours}) detailed in Section \ref{sec:irs}. 

\subsection{Imbalance Robust Softmax (IR-Softmax)}
\label{sec:irs}

\begin{figure*}[!htp]
\centering
\subfigure[Softmax,  balanced ]{
\includegraphics[width=0.31\linewidth]{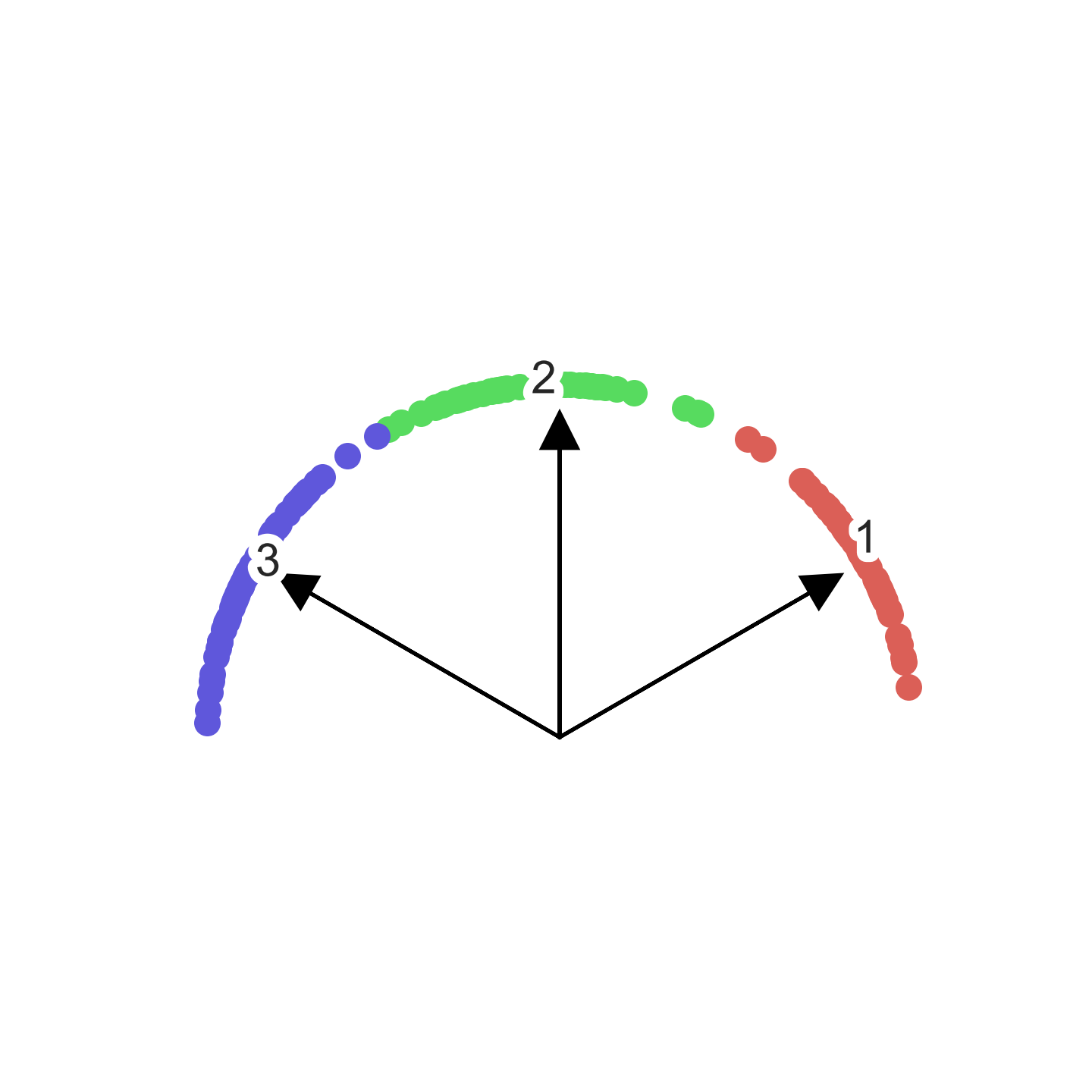}
\label{fig:softmax-dis}
}
\centering
\subfigure[A-Softmax,  imbalanced ]{
\includegraphics[width=0.31\linewidth]{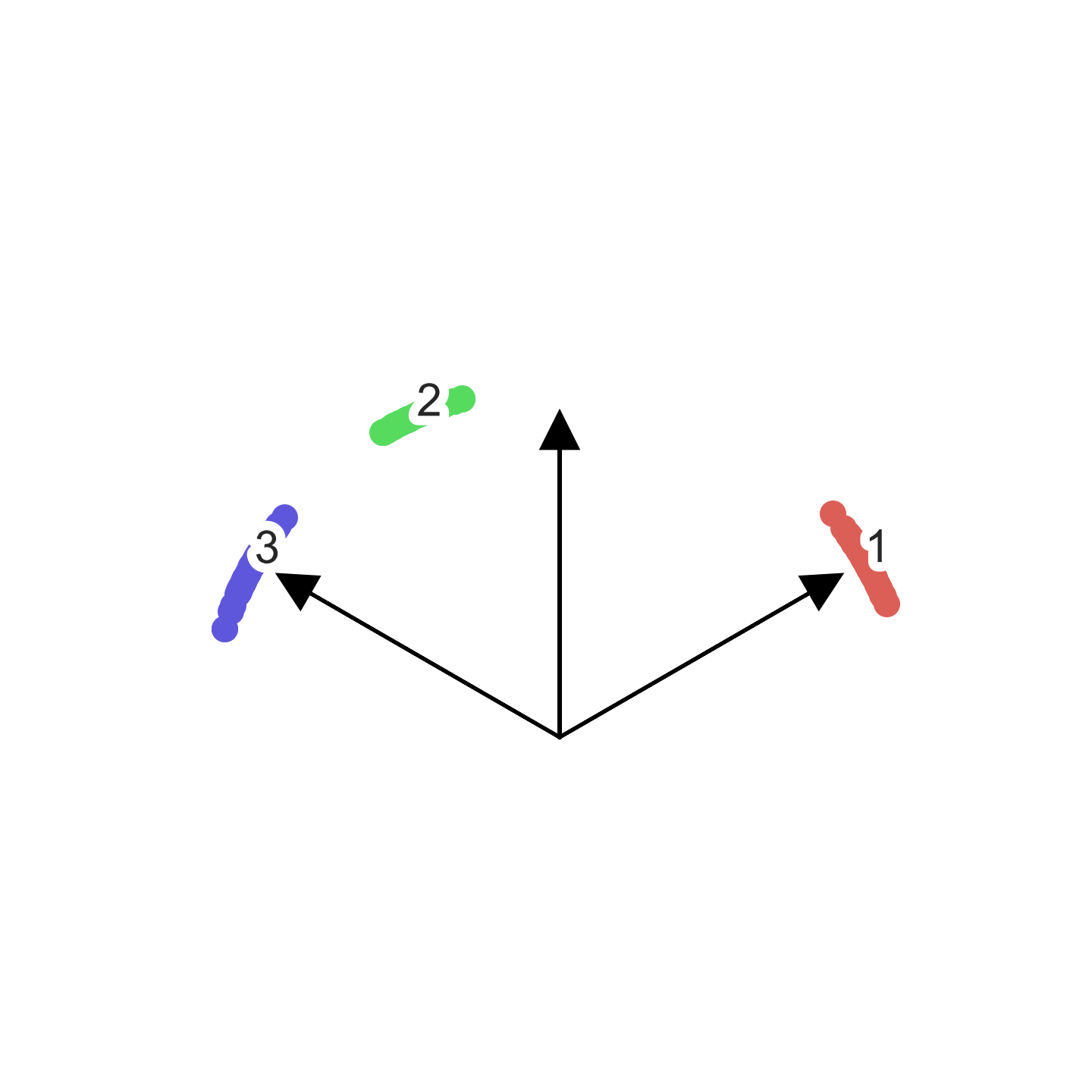}
\label{fig:sphere-dis}
}
\centering
\subfigure[IR-Softmax, imbalanced]{
\includegraphics[width=0.31\linewidth]{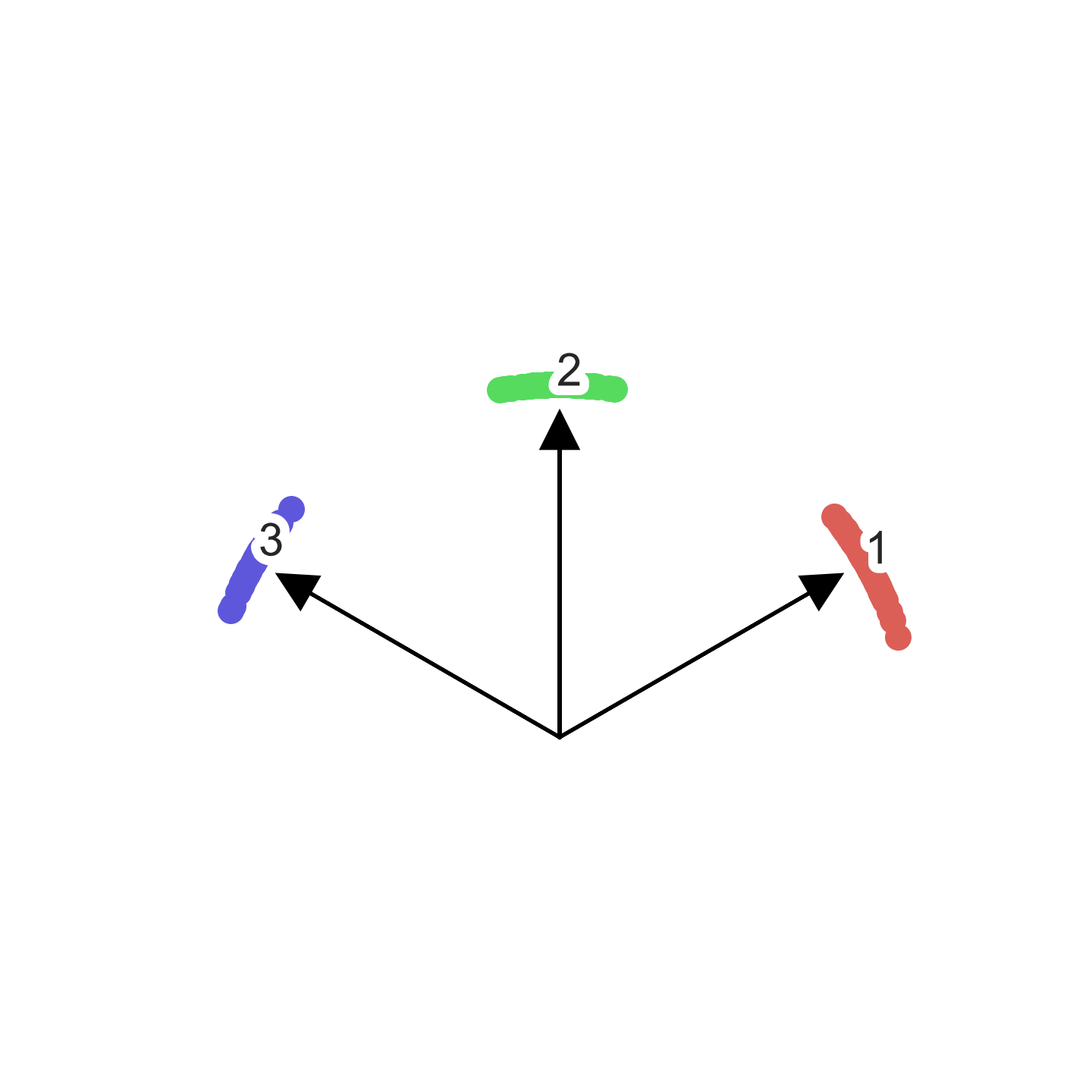}
\label{fig:target}
}
\caption{ Feature distributions in angular space: (a) separable features under balanced  data, feature center being close to its weight (the black arrow); (b) compact features under imbalanced data, and the feature center being far from its weight;  (c)  compact features under imbalanced data, 
the feature center being close to its weight.
}
\label{fig:fig2}
\end{figure*}

In this work, we aim to learn features which can   (i) improve the discriminative power of features in open-set protocol, and (ii) alleviate data imbalance problem. For (i), the desired open-set criterion is that the maximal intra-class distance is smaller than the minimal inter-class distance. However, softmax only maximises the the distance between weights rather then inter-class distance (Fig~.\ref{fig:softmax-dis}). Derived from softmax, A-Softmax \cite{liu2017sphereface}, however, only focuses on minimising intra-class distance, leading to compact features as shown in Fig.~\ref{fig:sphere-dis}. 
For (ii), data imbalance can  degrade the performance of deep CNNs. As analysed in Section~\ref{s:softmax}, in particular Eq. (\ref{equ:softmax-gradient}), the data imbalance can cause imbalanced gradient updates during optimisation: the strength of making the weights close to feature center (term 1 in Eq. (\ref{equ:softmax-gradient})) is much smaller than the strength of pushing away (term 2). This strength imbalance causes the weights being far away from their feature centers as shown in Fig. \ref{fig:sphere-dis}
, \ref{fig:Isoftmax} and \ref{fig:ILM}.

Based on the above analysis, to simultaneously solve the open-set  and data imbalance problems, two criteria corresponding these two problems are: 
1) minimising the intra-class distance by making the feature distribution with the same label more compact;
2) maximising the low bound of inter-class distance by making the center of features of each class being close enough  (ideally equal) to its  weights (usually the penultimate fully-connect layer). 
Criterion 1) can be achieved by learning compact features e.g. A-Softmax \cite{liu2017sphereface}  and AM-Softmax \cite{2018arXiv180105599W}. 
To our knowledge, we are the first to investigate  Criterion 2). 
To simultaneously achieve these two targets, we propose a novel framework, IR-Softmax\footnote[1]{code is available in  https://github.com/allenhaozhu/IR-Softmax}, which can achieve Criterion 1) by  incorporating itself into   discriminative softmaxs  e.g. A-Softmax \cite{liu2017sphereface}  and AM-Softmax \cite{2018arXiv180105599W}.

Now we detail the way of meeting Criterion 2). 
As the analysis in Section \ref{s:softmax}, we can find that the imbalanced data causes the weights (from penultimate fully-connected layer) being away from their class centers after training as shown in Fig. ~\ref{fig:Isoftmax}, leading to degraded classification performance. Based on Criterion 2), the key idea of IR-Softmax is \emph{setting the weights as  their corresponding class centers in the feature space}, naturally avoiding the shift between the weights and their centers. 

IR-Softmax is a unified framework which can be incorporated into softmax and its variants, leading to different forms of IR-Softmax. 
For classical softmax in Eq. (\ref{equ:softmax}), $f_j = \mathbf{w}_j^T \mathbf{x}_i + b_j$ is fed into softmax. In our IR-Softmax framework,  $f^{'}_j = (\mathbf{c}^{'}_j)^{T} \mathbf{x}_i + b_j$ replaces $f_j$, where $\mathbf{c}^{'}_j$, the center of features from class  $j$, is defined as:

\begin{equation}
\mathbf{c}_j' = \frac{1}{\sum_i^n 1\{ y_i = j\}}\sum_i^n\frac{1\{ y_i = j\} \mathbf{x}_i}{\|\mathbf{x}_i\|_2}
\label{equ:GC}
\end{equation}

Most importantly, in this work, we formulate two discriminative IR-Softmaxs derived from A-Softmax and AM-Softmax, respectively. To formulate A-Softmax  and AM-Softmax which both normalise the weight $w_j$ ($||\mathbf{w}_j||_2=1$), Eq. (\ref{equ:softmax}) can be modified as: 
\begin{equation}
\label{equ:oriIS}
J = - \left[ \sum_{i=1}^{m} \log \frac{\exp(\|\mathbf{x}_i\|\psi(\theta_{y_i}))}{\exp(\|\mathbf{x}_i\|\psi(\theta_{\mathbf{y}_i}))+\sum_{j\neq y_i}^K \exp(\|\mathbf{x}_i\|\cos(\theta_j))}\right]
\end{equation}
From   $f_j = \mathbf{w}_j^T \mathbf{x}_i + b_j$ (classical softmax) and $f^{'}_j = (\mathbf{c}^{'}_j)^{T} \mathbf{x}_i + b_j$ (IR-Softmax version),
we use $\mathbf{c}^{'}_j$ to replace $\mathbf{w}_j$. 
Similarly, we use 
$\mathbf{c}_j^T\mathbf{x}_i$ ($\mathbf{c}_j = \frac{\mathbf{c}_j'}{\|\mathbf{c}_j'\|_2}$) to replace $\|\mathbf{x}_i\|\cos(\theta_j)$ for both A-Softmax and AM-Softmax. Thus, our IR-Softmax version of Eq. (\ref{equ:oriIS}) is:
\begin{equation}
\label{equ:oriISE}
J = - \left[ \sum_{i=1}^{m} \log \frac{\exp(\|\mathbf{x}_i\|\psi(\theta_{y_i}))}{\exp(\|\mathbf{x}_i\|\psi(\theta_{y_i}))+\sum_{j\neq y_i}^K \exp(\mathbf{c}_j^T\mathbf{x}_i)}\right]
\end{equation}
For A-Softmax, 
\begin{equation}
\label{equ:sphereface}
\psi(\theta_{y_i}) = ((-1)^k\cos(m\theta_{y_i})-2k), \theta_{y_i}\in[\frac{k\pi}{m},\frac{(k+1)\pi}{m}]
\end{equation}
where  $k\in[0,m-1]$ and $m\geq1$ is an integer that controls the size of angular margin.
For original A-Softmax, $\theta_{y_i} (0 \leq \theta_i \leq \pi)$ is the angle between $\mathbf{w}_i$ and  $\mathbf{x}_i$. Note that, in our IR-Softmax, $\theta_{y_i} $ is \emph{the angle between the  $\mathbf{c}_{y_i}$ and $\mathbf{x}_i$}, where $\mathbf{c}_{y_i}$ is the center of class $i$.
For AM-Softmax, 
\begin{equation}
\psi(\theta_{y_i}) = \cos(\theta_{y_i}+\alpha)
\end{equation}

\begin{equation}
\psi(\theta_{y_i}) = \cos(m_1 \theta_{y_i}-m_2)-m_3
\end{equation}
where $\theta_{y_i} (0 \leq \theta_{y_i} \leq \pi)$  of the original AM-Softmax is the angle between $\mathbf{w}_i$ and $\mathbf{x}_i$.
Note that  $\theta_{y_i}$ of IR-Softmax is \emph{the angle between the  $\mathbf{c}_{y_i}$ and $\mathbf{x}_i$}. 

Now we can summarise the difference between IR-Softmax and softmax (and its variants). 
First, the weight $\mathbf{w}_i$ of softmax is updated via back-propagation, however,  $\mathbf{c}_i$ of IR-Softmax can be computed directly from Eq.~(\ref{equ:GC}). 
Second, the update of   $\mathbf{w}_i$  depends on samples of class $i$ and  samples from other classes as shown in Eq.~(\ref{equ:softmax-gradient}). In contrast, the update of $\mathbf{c}_i$ of IR-Softmax only depends on the samples from class $i$, effectively avoiding the influence of data imbalance.

In practice, it is impossible to use all samples to calculate the centers as shown in Eq.(\ref{equ:GC}).
%
We have tried three different updating strategies for feature centres.  i.replacing the weight with an instance feature (which makes the proposed method like docFace~\cite{shi2019docface+}). ii.memory bank: estimate the centre with last few (in a fixed window) samples in the same class (without BP). iii.a loss function to estimate centres on the unit sphere for different classes.
The disadvantage of the first solution is that an additional softmax is necessary in case the convergence is slightly slow and unstable. The second solution relieves the unstable issue but no improvement in performance. The third method is equal to adding a new term $\| \mathbf{c}_i - \frac{\mathbf{x}_i}{\|\mathbf{x}_i\|_2}\|_2^2, s.t. \|\mathbf{c}_i\|=1$ to Eq.\ref{equ:oriIS} and use the corresponding BP process to update feature centres (weights in Eq.5) rather than Eq.\ref{equ:oriIS}. After that $\ell2$ normalization is used on feature centres to make sure new $c_i$ on the unit sphere. We select the third one in our experiments because it works better than other two approaches.

\subsection{Relation to Metric Learning}
N-pairs loss \cite{sohn2016improved} enforces softmax cross-entropy loss among the pairwise similarity  in the \emph{mini-batch}.
\begin{equation}
\label{equ:n-pair-mc}
E = \frac{-1}{|P|}\sum_{(i,j)\in P}\log\frac{\exp(S_{i,j})}{\exp(S_{i,j})+\sum_{k:y_k\ne y_j}\exp(S_{i,k})}
\end{equation}
where $S_{i,j} = f(\mathbf{x}_i,\mathbf{\Theta})^Tf(\mathbf{x}_j,\mathbf{\Theta})$ represents the inner product between two embeddings.
and $|P|$ indicates the number of positive pairs ($i$,$j$). 
Comparing  Eq.~(\ref{equ:oriIS}) and ~(\ref{equ:n-pair-mc}), we can see our  method can be viewed as a special form of N-pair loss. However, there are two main differences: (1) Unlike softmax embedded in N-pairs, we employ A-Softmax \cite{liu2017sphereface} and AM-Softmax~\cite{2018arXiv180105599W} to improve the discriminability of features. (2) The size of \emph{mini-batch} (where N-pair works) limits {the number of negative samples}. In practice, it is usually difficult to make mini-batch more than  256 due to the memory limitation of GPU. In contrast, our method alleviates the problem by caching historical features. The positive and negative samples are equal to the number of categories. 

\section{Experiments}
\label{sec:exp}
In this section, we first describe the  experimental settings. We then evaluate our method on two different tasks, face recognition (FR) and person re-identification (re-ID), against four different benchmarks. For FR, we use CASIA-WebFace \cite{DBLP:journals/corr/YiLLL14a} as training set and evaluate our method on LFW \cite{huang2007labeled} and MegaFace \cite{kemelmacher2016megaface}. For re-ID, we evaluate on the Market-1501 \cite{zheng2015scalable} and Duke \cite{ristani2016performance} datasets.

\subsection{Face Verification}
All the faces  and their landmarks are detected by MTCNN \cite{zhang2016joint}. We use the detected 5 landmarks (two eyes, nose and two mouth corners) to perform similarity transformation. When the detection fails, we simply discard the image if it is in the training set, but use the provided landmarks if in the test set. 

We use the publicly available training dataset CASIA-WebFace \cite{DBLP:journals/corr/YiLLL14a} (excluding the images of 59 identities appearing in testing sets \cite{2018arXiv180105599W}) to train our CNN models. CASIA-WebFace has 494,414 face images belonging to 10,575 (in fact, 10,516 after removing) different individuals. As shown in Fig.~\ref{fig:imbalance-dis}, CASIA-WebFace is an imbalance dataset. Some identities  have very few images (e.g, only one image), while some  have more than 300 images.
These face images are horizontally flipped for data augmentation in the training process. Note that the number of samples in training set (0.49M) is relatively small compared to other private datasets used in DeepFace \cite{sermanet2013overfeat} (4M), VGGFace \cite{parkhi2015deep} (2M) and FaceNet \cite{schroff2015facenet} (200M). In the testing process, we extract the deep features from the output of the FC1 layer and do not employ any pre-processing (such as PCA and  flipped features).  The cosine distance between two features is applied. A nearest neighbor classifier and thresholding are used for face identification and verification, respectively.

To make fair comparison, we use two widely used CNN architectures for face recognition: 9-layer Light CNN \cite{wu2015lightened} and 20-layer ResNet-20 \cite{liu2017sphereface}. 
Note that the faces are cropped to two different sizes (128x128 and 112x96) to fit the setting in \cite{wu2015lightened} and \cite{liu2017sphereface} respectively. In the training process, our IR-Softmax is appended after the feature layer, i.e. the second last inner-product layer. The networks  are trained in an end-to-end way. 

For simplicity, we denote IR-Softmax (A) as our IR-Softmax instance derived from  A-Softmax, and IR-Softmax (AM) from AM-Softmax in the whole experiment section.

\subsubsection{LFW}
The LFW dataset \cite{huang2007labeled} contains 13,233 images from 5,749 identities, with large variations in pose, expression and illumination. All the images are collected from the internet. We evaluate our methods on two protocols: (1) official protocol \cite{huang2007labeled} and (2) BLUFR protocol \cite{liao2014benchmark}. 
For (1),  LFW is divided into
10 predefined splits for cross validation. We follow the standard `Unrestricted, Labeled Outside Data' protocol. Because the performance of face recognition is almost saturated on this protocol, researchers propose a more challenging  BLUFR protocol \cite{liao2014benchmark}. For (2),
BLUFR utilises all 13,233 images to evaluate the performance in the open-set setting. The Verification Rate (VR) at False Accepted Rate (FAR) 0.1\% (VR@FAR=0.1\%) and Detection and Identification rate (DIR) at FAR 1\% (DIR@FAR=1\%) are reported under BLUFR. It is noteworthy that  not only three  identities exist in both CASIA-Webface \cite{DBLP:journals/corr/YiLLL14a,2018arXiv180105599W} and LFW \cite{huang2007labeled}. We removed them according to \cite{2018arXiv180105599W} during training to build a complete open-set validation. 

\paragraph{LFW Official Protocol, Light CNN}
As shown in Table~\ref{tab:Light}, the performance is evaluated by six methods. The proposed IR-Softmax(A) and IR-Softmax(AM) greatly outperform their original versions (A-Softmax and AM-Softmax). 
Compared with the baseline (i.e. Softmax), IR-Softmax (A)  improves the verification accuracy from 97.15\% to 98.38\%, and IR-Softmax (AM)  from 97.15\% to 98.63\%.

\paragraph{LFW Official Protocol, ResNet-20}
The evaluation results  of ResNet-20 
are listed in Table~\ref{tab:RES}. Other state-of-the-art results of A-Softmax and AM-Softmax using ResNet-20 are also presented for comparison. Compared with the baseline (i.e. Softmax), IR-Softmax (A) loss  improves the verification accuracy from 97.08\% to 99.23\% on LFW. From the results, we can see that the proposed methods IR-Softmax(A) and IR-Softmax(AM) can outperform the corresponding original versions.

\paragraph{BLUFR, Light CNN}
From Table~\ref{tab:Light}, we can observe that the proposed method significantly outperforms the other methods \cite{liu2016large,liu2017sphereface}. Specifically, IR-Softmax(A) beats the softmax baseline (which we finetune our model from), and improves the VR@FAR=0.1\% from 83.32\% to 94.61\%,  while DIR@FAR=1\% from 60.64\% to 75.12\%. Both versions of IR-Softmax are able to outperform their counterparts. It means that the proposed method can significantly enhance the discriminability of deeply learned features in the open-set protocol, demonstrating the effectiveness of the proposed method.

\paragraph{BLUFR, ResNet-20} 
{
Since ResNet-20 models \cite{liu2017sphereface,DBLP:journals/corr/WangXCY17}  are also widely used for face recognition, we  make comparisons based on ResNet-20  in Table~\ref{tab:RES}. 
}
IR-Softmax with ResNet-20  keeps the similar superiority compared with other models in the BLUFR protocol of LFW. Note that our approach is better than range loss, which is proposed to solve the problem of data imbalance in face recognition. Though range loss uses a larger training set (MS-celeb \cite{guo2016msceleb}) and a deeper network (ResNet-50), our method still outperforms it with VR@FAR=0.1\% from 93.72\% to 97.08\% (IR-Softmax(A)) or 98.09\% (IR-Softmax(AM)) while DIR@FAR=1\% from 71.11\% to 81.52\% (IR-Softmax(A)) or 85.00\% (IR-Softmax(AM)). 



\begin{table}[t]
\centering
\caption{Performance on ResNet with various loss functions. CenterLoss, NormFace model and sphereface model are provided by authors. NormFace and CenterLoss use ResNet-28 like \cite{wen2016discriminative}, another two methods use ResNet-20 \cite{liu2017sphereface}. }
\begin{tabular}{ | c | c | c | c | c | c | }
\hline
	loss function & LFW~\cite{huang2007labeled}& BLUFR \cite{liao2014benchmark}& BLUFR \cite{liao2014benchmark}& MegaFace \cite{kemelmacher2016megaface}  & MegaFace \cite{kemelmacher2016megaface} \\
    & 6000 pairs & VR@FAR=0.1\% &  DIR@FAR=1\% & rank1@1e-6 & VR@FAR=1e-6\\ \hline
	Softmax & 97.08\% & 78.26\% & 50.85\% & 45.26\% & 50.12\% \\ \hline
	CenterLoss \cite{wen2016discriminative} & 99.00\% & 94.50\% & 65.46\% & 63.38\% & 75.68\% \\ \hline
	NormFace \cite{DBLP:journals/corr/WangXCY17} & 98.98\% & 96.16\% & 75.22\% & 65.03\% & 75.88\% \\ \hline
	A-Softmax \cite{liu2017sphereface}& 99.08\% & 96.58\% & 79.97\% & 67.41\% & 78.19\% \\ \hline
	IR-Softmax(A) & 99.23\% & 97.08\% & 81.52\% & 69.48\% & 80.32\% \\ \hline
	AM-Softmax \cite{2018arXiv180105599W} & 98.98\% & 97.69\% & 84.82\% & 72.47\% & 84.44\% \\ \hline
    IR-Softmax(AM) & 99.21\% & 98.09\% & 85.00\% & 75.28\% & 85.67\% \\ \hline
\end{tabular}
\label{tab:RES}
\end{table}

\begin{table}[t]
\centering
\caption{Performance on Lighten CNN with various loss functions. All Results are derived under the same settings used in \cite{wu2015lightened}.}
\begin{tabular}{ | c | c | c | c | c | c | }
\hline
	loss function & LFW~\cite{huang2007labeled}& BLUFR \cite{liao2014benchmark}& BLUFR \cite{liao2014benchmark}& MegaFace \cite{kemelmacher2016megaface}  & MegaFace \cite{kemelmacher2016megaface} \\
    & 6000 pairs & VR@FAR=0.1\% &  DIR@FAR=1\% & rank1@1e-6 & VR@FAR=1e-6\\ \hline
	Softmax & 97.15\% & 83.32\% & 60.64\% & 47.31\% & 54.86\% \\ \hline
	Large-Margin \cite{liu2016large} & 98.35\% & 91.62\% & 64.76\% & 59.03\% & 70.57\% \\ \hline
	A-Softmax \cite{liu2017sphereface} & 98.20\% & 91.16\% & 66.55\% & 54.87\% & 60.75\% \\ \hline
	IR-Softmax(A) & 98.38\% & 94.61\% & 75.12\% & 64.71\% & 75.94\% \\ \hline
	AM-Softmax \cite{2018arXiv180105599W} & 98.58\% & 94.67\% & 72.80\% & 65.33\% & 78.76\% \\ \hline
    IR-Softmax(AM) & 98.63\% & 95.36\% & 79.92\% & 66.71\% & 78.83\% \\ \hline
\end{tabular}
\label{tab:Light}
\end{table}

\subsubsection{MegaFace}
One of the most challenging datasets for face recognition  is MegaFace \cite{kemelmacher2016megaface}. The MegaFace dataset contains a gallery set and a probe set. The gallery set contains more than 1 million images from 690K identities; The probe set consists of two existing datasets: Facescrub \cite{ng2014data} and FGNet. MegaFace has multiple testing scenarios including identification, verification and pose-invariance under two protocols i.e. large or small training sets. The training set is considered small if it is less than 0.5M. We evaluate our IR-Softmax under the small training set protocol. 
\paragraph{Lighten-CNN}  Table~\ref{tab:Light} shows that our IR-Softmax(A) outperforms A-Softmax result by a margin (almost 10\% for rank-1 identification rate and 15\% for VR at 1e-6 FAR) on the small training dataset protocol while IR-Softmax(AM) outperforms AM-Softmax result by a margin (7\% for rank-1 identification rate and 1.4\% for VR at 1e-6 FAR).
Compared to the softmax baseline, our method performs significantly better: 15\% from IR-Softmax (A) and  19\% IR-Softmax(AM) for  identification,   21\% from IR-Softmax (A) and 24\% from IR-Softmax (AM) for  verification.

\paragraph{ResNet-20}  Table~\ref{tab:RES} shows that our IR-Softmax (A) outperforms A-Softmax result by a margin (almost 2\% for rank-1 identification rate and 2\% for VR at 1e-6 FAR) on the small training dataset protocol while IR-Softmax(AM) outperforms AM-Softmax result by a margin ( almost 3\% for rank-1 identification rate and 1.2 \% for VR at 1e-6 FAR).
Compared to the softmax baseline, our method performs significantly better: 24\% from IR-Softmax (A) and 30\% from IR-Softmax (AM) for  identification,   30\% from IR-Softmax (A) and 35\% IR-Softmax (AM) for  verification.

Note that the performance of any testing methods on Megaface is intimately linked to the quality of face alignment. Thus we do not compared with other methods with different alignments. The results in Table~\ref{tab:Light} are therefore computed under the same setting of face alignment and are directly comparable.
These results  demonstrate that our IR-Softmax is well designed for open-set face recognition especially when the training set is imbalanced. One can also see that,  smaller intra-class distance is not the only important issue for learning features, but larger and evenly inter-class angular margin can significantly improve face recognition performance.

\subsection{Person Re-identification}


For the evaluation of re-ID, we focus on two well-known re-ID databases: Market-1501 \cite{zheng2015scalable} and DUKE \cite{ristani2016performance} datasets. As shown in Fig.~\ref{fig:imbalance-dis}, we demonstrate the distribution of market-1501 database. Although there are no identities with more than 100 images like WebFace dataset, the number of images per person  ranges from 5 to 80. The DUKE \cite{ristani2016performance} also has the similar imbalance pattern. 
We use the standard evaluation metrics for both datasets, namely the mean average precision score (mAP) and the cumulative matching curve (CMC) at rank-1. We follow common practice by using random crops and random horizontal flips during training. Specifically, we resize all images to  $256\times128$, of which we take random crops of size $224\times112$. Many methods for re-ID rely on pre-trained models (e.g. ResNet). Indeed, these models usually lead to impressive results. However, pre-trained models reduce the flexibility to make task-specific changes in a network. For example, some application scenarios need compact models rather than large ones pre-trained on Imagenet. Our method clearly suggests that it is also possible to learn deep  models from scratch and achieve state-of-the-art performance. We use a  Lighten CNN \cite{wu2015lightened} based on the ResNet Architecture, which is faster than the current ResNet-50 used by many works \cite{DBLP:journals/corr/SunZDW17}.
%
Compared with other methods, we do not use the corresponding pretrained models in ImageNet for finetuning. Thus we use the softmax to train a baseline model with the re-ID dataset directly. And other methods (e.g. large-margin and the proposed method) employ the baseline model as the pre-trained model and finetune this model further. 

\subsubsection{Market-1501}
The Market-1501 dataset contains 1,501 identities, 19,732 gallery images and 12,936 training images captured by 6 cameras. All the bounding boxes are generated by the DPM detector \cite{felzenszwalb2008discriminatively}. The dataset uses both single and multi-query evaluation, we report the results for both. Table~\ref{tab:market} compares our IR-Softmax (A) to other approaches. For Market-1501, the improvements achieved by IR-Softmax are significant: (1) Compared with softmax, the Rank-1 accuracy  rises from 81.47\% to 91.87\%, and the mAP  from 57.42\% to 76.72\% in the setting of single query; (2) In the setting of multi query, the Rank-1 accuracy  rises from 86.40\% to 94.33\%, and the mAP  from 65.97\% to 82.22\%. IR-Softmax (A) significantly outperforms not only the softmax baseline but also other state-of-the-art methods \cite{zheng2017unlabeled}.

\subsubsection{DukeMTMC-reID}
The DukeMTMC-reID dataset is collected via 8 cameras and used for cross-camera tracking (handover). Table~\ref{tab:duke} compares our IR-Softmax to other approaches. For DukeMTMC-reID, IR-Softmax(A) works much better than softmax: the Rank-1 accuracy: 76.84\% vs 61.98\%, and the mAP 57.47\% vs 41.17\%. Beyond that, the imbalance robust softmax also outperforms other state-of-the-art methods\cite{DBLP:journals/corr/SunZDW17}.

\begin{table}[t]
\centering
\caption{Comparison with the state-of-the-art methods on the Market-1501 dataset. The rank-1 accuracy and mAP on single and multiple query are reported respectively.}
\begin{tabular}{|c|cc|cc|}
\hline
\multirow{2}{*}{Method} & \multicolumn{2}{c|}{Single Query}&\multicolumn{2}{c|}{Multi. Query} \\
 & rank-1 & mAP & rank-1 & mAP \\
 \hline
 \hline
BoW +KISSME \cite{zheng2015scalable} & 44.42 & 20.76 & - & - \\
MR CNN \cite{DBLP:journals/corr/UstinovaGL15}& 45.58 & 26.11 & 56.59 & 32.26 \\
DSN \cite{DBLP:conf/cvpr/ZhangXG16}& 55.43 & 29.87 & 71.56 & 46.03 \\
Gate Reid \cite{DBLP:conf/eccv/VariorHW16}& 65.88 & 39.55 & 76.04 & 48.45 \\
SOMAnet \cite{DBLP:journals/corr/BarbosaCCRT17}& 73.87 & 47.89 & 81.29 & 56.98 \\
DeepTransfer \cite{DBLP:journals/corr/GengWXT16}& 83.70 & 65.50 & \textbf{89.60} & 73.80 \\
Basel+LSRO \cite{zheng2017unlabeled}& \textbf{83.97} & \textbf{66.07} & 88.42 & \textbf{76.10} \\
SVDNet \cite{DBLP:journals/corr/SunZDW17}& 82.30 & 62.10 & - & - \\
\hline
\hline
Softmax   & 81.47 & 57.42 & 86.40 & 65.97\\
Large-margin \cite{liu2016large} & 90.08 & 72.22 & 92.75 & 78.79 \\
IR-Softmax(A)& \textbf{91.87}& \textbf{76.72} & \textbf{94.33} & \textbf{82.88} \\
\hline
\end{tabular}
\label{tab:market}
\end{table}

\begin{table}[t]
\centering
\caption{Comparison with state-of-the-art methods on DukeMTMC-reID. Rank-1 accuracy and mAP are reported.}
\begin{tabular}{|c|c|c|}
\hline
Method & Rank-1 (\%) & mAP (\%)\\
\hline
\hline
BoW + KISSME \cite{zheng2015scalable}& 25.13 & 12.17 \\
LOMO + XQDA \cite{DBLP:conf/cvpr/LiaoHZL15}& 30.75 & 17.04 \\
Basel + LSRO \cite{zheng2017unlabeled}& 67.68 & 47.13 \\
ACRN \cite{schumann2017person}& 72.58 & 51.96 \\
PAN \cite{DBLP:journals/corr/ZhengZY17aa}& 71.59 & 51.51 \\
SVDNet \cite{DBLP:journals/corr/SunZDW17}& \textbf{76.70} & \textbf{56.80} \\
\hline
\hline
Softmax & 61.98 & 41.17  \\
Large-margin \cite{liu2016large} & 75.58 & 56.25 \\ 
IR-Softmax(A) &\textbf{76.84} & \textbf{57.47}\\
\hline
\end{tabular}
\label{tab:duke}
\end{table}

\section{Conclusion}
In this paper, we investigated thoroughly the potential effects of data imbalance on the deep embedding learning and proposed a new framework,  Imbalance Robust Softmax (IR-Softmax). IR-Softmax can simultaneously solve the open-set problem and reduce the influence of data imbalance. 
Extensive experiments on FR and re-ID are conducted, and the results show the effectiveness of IR-Softmax. In Future work, we plan to extend this framework to more softmax based methods and other applications like few-shot learning.



\bibliographystyle{splncs}
\bibliography{egbib}

\end{document}